\documentclass[10pt]{article}
\usepackage{epsfig}
\usepackage{supertabular}
\usepackage{natbib}
\begin{document}

\title{Constructing word similarities in Meroitic as an aid to decipherment}
\maketitle
\author{Reginald D. Smith \\ Bouchet-Franklin Institute \\ P.O. Box 10051 \\ Rochester,
NY 14610, USA \\ \texttt{rsmith@bouchet-franklin.org}}

\begin{abstract}
Meroitic is the still undeciphered language of the ancient
civilization of Kush. Over the years, various techniques for
decipherment such as finding a bilingual text or cognates from
modern or other ancient languages in the Sudan and surrounding areas
has not been successful. Using techniques borrowed from information
theory and natural language statistics, similar words are paired and
attempts are made to use currently defined words to extract at least
partial meaning from unknown words.
\end{abstract}

This paper addresses a technique using a combination of known words
and techniques from information theory to try to decipher the
meanings of additional words in the extinct and undeciphered
language, Meroitic. First, I will give a short history of the
language and the problems translating it and next describe the
statistical techniques and their results and implications.

\section{A Short History of Meroitic \citep{torok, lobban}}

\begin{figure}
    \centering
    \includegraphics[height=7in, width=5in]{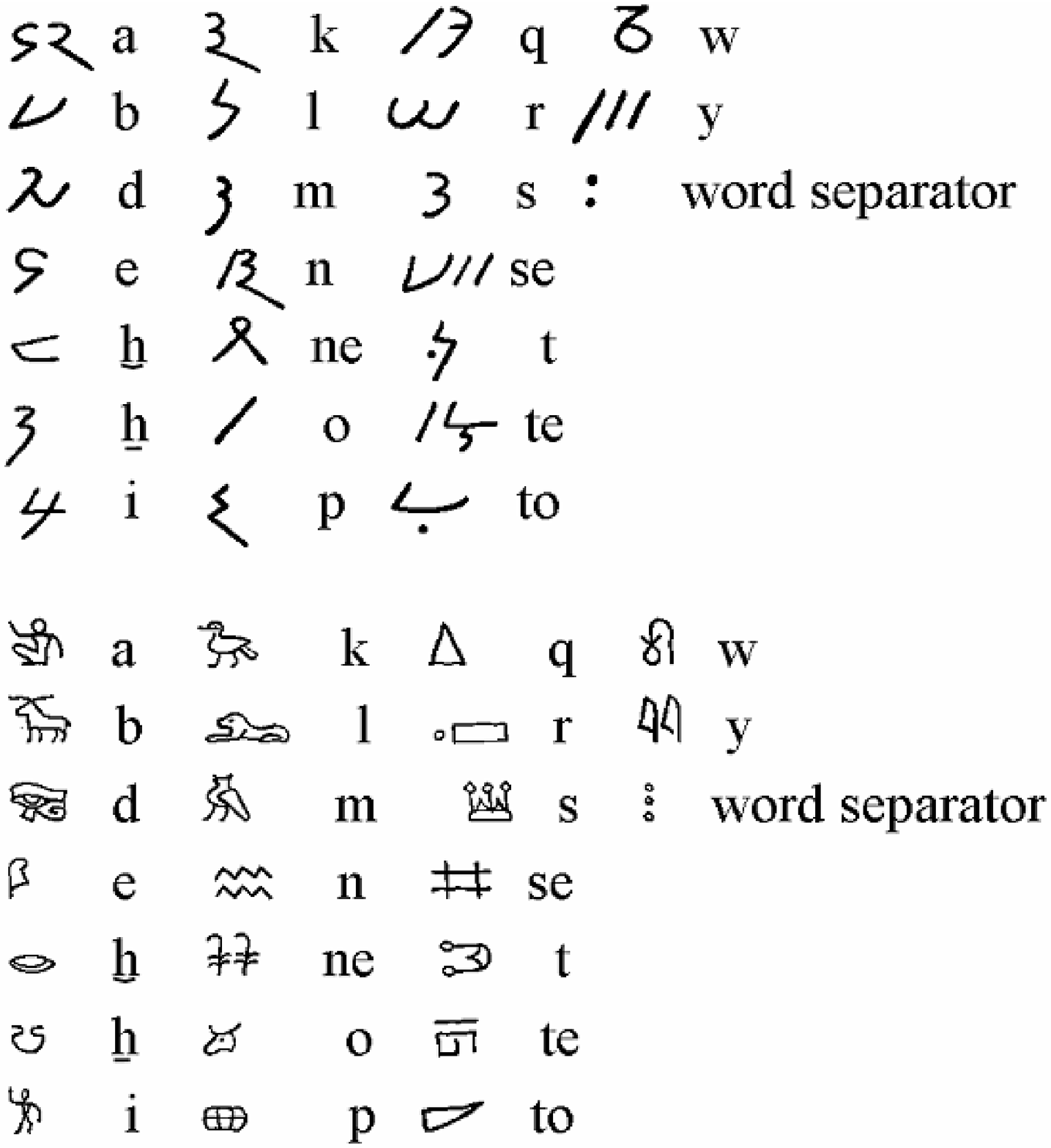}

\caption{Meroitic Cursive and Hieroglyphic words and their
transliterations. Taken from the latest font set for Meroitic
Hieroglyphic and Cursive characters developed by the Meroitic
scholars Claude Carrier, Claude Rilly, Aminata Sackho-Autissier, and
Olivier Cabon. Web Address:
http://www.egypt.edu/etaussi/informatique/meroitique/meroitique01.htm}
\label{symbols}
\end{figure}

Meroitic was the written language of the ancient civilization of
Kush, located for centuries in what is now the Northern Sudan. The
word 'Meroitic' derives from the name of the city Meroë, which was
located on the East bank of the Nile south of where the Atbara River
flows off to the east. It is the second oldest written language in
Africa after Egyptian hieroglyphs. It is a phonetic language with
both a hieroglyph form using some adopted Egyptian hieroglyphs and a
cursive form similar to Egyptian Demotic writing. The language had
one innovation uncommon in ancient written languages such as
Egyptian hieroglyphics or Greek in that there was a word separator,
similar in function to spaces in modern scripts, that looks similar
to a colon (see Figure \ref{symbols}). Meroitic was employed
starting the 2nd century BC and was continuously used until the fall
of Meroë in the mid 4th century AD.

The script was rediscovered in the 19th and 20th centuries as
Western archaeologists began investigating the ancient ruins in the
Sudan. The first substantial progress in deciphering Meroitic came
around 1909 when British archaeologist Francis Llewellyn Griffith
was able to use a bark stand which had the names of Meroitic rulers
in Meroitic and Egyptian hieroglyphs. The Meroitic hieroglyphs were
then corresponded to the Meroitic cursive script and it was then
possible to transliterate Meroitic (see Figure \ref{symbols}). Some
vocabulary was later deciphered by scholars including loan words
from Egyptian, gods, names, honorifics, and common items. However,
the language remains largely undeciphered. The greatest hope for
decipherment, a Rosetta stone type of tablet containing writing in
Meroitic and a known language such as Egyptian, Greek, Latin, or
Axumite, has yet to be found. Further confounding research is the
confusion regarding which language family Meroitic belongs to.
Cognate analysis has proceeded extremely slowly since it is disputed
to which language family Meroitic properly belongs. Recent work by
\citep{rilly} has suggested that Meroitic belongs to the North
Eastern Sudanic family, however, full decipherment is still elusive.

\section{Past Statistical and Mathematical Work on Meroitic}

Meroitic was one of the earliest ancient languages to be
investigated using computers
\citep{leclant,heyler,heyler2,ouellette}. Much of this work was
dedicated to creating an alphabetical index of Meroitic and also
comparing Meroitic words to possible cognates in Nubian or other
known ancient and modern languages from the region.

In \citep{smith}, many of the longest texts were analyzed by ranking
words according to frequencies to verify whether the current texts
we have follow the mathematical relation Zipf's Law where the word
frequencies $f$ vary with the rank $z$ according to the relation

\begin{equation}
f_{z} = \frac{C}{z^{\alpha}}, z=1,2,3,…n
\label{zipf}
\end{equation}

where $\alpha \approx 1$. In analyzing the Meroitic texts, though
many did not fit the strict criterion of $\alpha \approx 1$, the
frequency-rank distribution followed the behavior of a truncated
power law distribution whose exact parameters varied by text. Some
texts such as the long stela REM 1003 (REM is a text designation
that stands for R\'{e}pertoire d'\'{e}pigraphie m\'{e}ro\"{i}tique,
the most comprehensive catalogue of Meroitic texts) more closely fit
Zipf's Law though. From these results, without knowing the meaning
of the text it is clear that the statistical variations and
occurrences of words in the Meroitic texts in our possession are not
surprising and mirror those of other human languages. Though this
may seem a trivial property at first glance, it gives us the hope of
using more advanced statistical techniques to help tease some of the
meaning from the unknown portions of the language.

\section{Introduction to Statistical Techniques}

At the outset, I acknowledge that no language has ever been fully
deciphered using purely statistical or mathematical techniques and I
am not proposing that Meroitic will be completely understood using
these tools. In particular, many of the subtleties of human
semantics and syntax are irregular or do not follow a consistent
pattern that statistics is usually excellent at analyzing. What this
paper will attempt to do is not claim to derive the meaning, a
loaded concept in the study of linguistics, of a word but rather
find words which are used very similarly in the text. When two words
are used very similarly with one of the words being known, we can
hope to possibly infer what the other word in the pair means. In
linguistics, the hypothesis that words that appear in similar
contexts have similar semantics is known as the Distributional
Hypothesis \citep{harris,harris2}.

Similarity, which will be explained in more technical detail below,
will be defined by looking at whether two different words share
similar word neighbors for a distance of one or two words away. The
steps in analyzing the similarity are five-fold. First, I combined
several long Meroitic texts into one giant corpus. I separated out
some common bound morphemes to help better identify particular
words. Second, I used a computer program in Python to create three
matrices: one showing the relative frequency of each word, one
showing the frequency of a given word pair, WORD1:WORD2 for any
combination of the distinct words in the text for a word distance of
one, and a final array with word pair frequencies for a word
distance of two. Third, for all possible pairs of different words in
the texts, I used the frequency arrays to find the mutual
information between every distinct word pair. I created separate
arrays of the mutual information metric for the mutual information
based on word distance one and mutual information based on word
distance two and then calculated a blended mutual information based
on weightings of the one and two word distance mutual information.
Fourth, using the blended mutual information array, I used a
similarity metric to find similarity between words based on if they
had similar mutual information for the other words in the texts.
Finally, I compared the results for high similarity word pairs to
what is known about Meroitic words. A spanning tree graphically
showing the relationship between words was also aided to clarify the
similarity relationships.
\subsection{Step 1}

The long stelae texts REM 1001, REM 1003, and REM 1044A-D were
combined into one corpus separated by a character XXXX between the
beginning and end of each text. The XXXX made sure that the last
word of one text and the first word of another are not accidentally
matched for either a distance one or two word pair. In addition, as
in \citep{smith} several common and recognized bound morphemes were
separated from the words by the word separator character so they
would be treated as separate words. Many Meroitic verbs, as well, as
some nouns, have suffixes which contain grammatical meaning. For
example, it is known that the suffix \emph{telowi} or \emph{teli} is
appended to the name of a place, such as a city, to indicate that
the subject of the sentence was affiliated with this place. There is
also an extremely common suffix \emph{lowi} (``he/she/it is'') or
\emph{li} (``the'') that is appended to nouns that may denote the
noun as an indirect object in the sentence. Their definitions are
still tenuous, however, these bound morphemes are very common and
were separated into independent words for the second Zipf plot. The
six bound morphemes separated out were ``\emph{qo}'', ``\emph{lo}'',
``\emph{li}'', ``\emph{te}'', ``\emph{lebkwi}'', ``\emph{mhe}''.
They were separated in the manner:

\begin{table*}[!t] \vspace{1.5ex}

\begin{tabular}{c c}

\emph{qo}$\rightarrow$ separated out to
``\emph{qo}''&\emph{lw}$\rightarrow$ separated out to ``\emph{lw}'' \\
\emph{atomhe}$\rightarrow$\emph{ato} and
\emph{mhe}&\emph{lo}$\rightarrow$ separated out to ``\emph{lo}''\\
\emph{telowi}$\rightarrow$ \emph{te} and \emph{lo} and
\emph{wi}&\emph{atmhe}$\rightarrow$ \emph{at} and \emph{mhe}\\
\emph{li}$\rightarrow$ separated out to
``\emph{li}''&\emph{teli}$\rightarrow$ \emph{te} and \emph{li}\\
\emph{qowi}$\rightarrow$ \emph{qo} and
\emph{wi}&\emph{lowi}$\rightarrow$ \emph{lo} and \emph{wi}\\
\emph{lebkwi}$\rightarrow$ \emph{lebk} and \emph{wi}
\end{tabular}
\caption{Meroitic bound morpheme separations}
\end{table*}

\subsection{Step 2}

The word frequency arrays were created as follows. First, a
normalized frequency of each different word in the text was
calculated ranging between 0 and 1 where the total frequency of a
word divided by the total number of words in a text defines the word
frequency. Next to understand word pair frequency, imagine a string
of words separated by the colon-like word separator character,
A:B:C. B/C and A/B are distance one neighbors and A/C are distance
two neighbors. This is repeated for all words throughout the text.
The frequency of a word pair is the number of occurrences of that
pair divided by the total number of word pairs in the text.

\subsection{Step 3}

Here the procedure becomes more complicated and theoretical so the
appropriate background is necessary. Many statistical natural
language methods for analyzing corpuses such as hidden Markov models
(HMM) or neural networks require ``training'' with a tagged corpus
that emphasizes parts of speech, grammar, etc. Since these are
mostly unknown for Meroitic, we are forced to rely on techniques
that make no \emph{a priori} assumptions about the language syntax
or word relationships.

Two relatively similar approaches relying on the Distributional
Hypothesis were used in \citep{lankhorst}and \citep{lin} in
combination with genetic algorithms and similarity measures
respectively to find relationships between words based on their
distributions within a text. In \citep{lankhorst}, a fixed number of
categories is created and each word is randomly assigned a category.
The mutual information among words in each category is measured and
the categories are altered using a genetic algorithm with mutual
information as the fitness. A maximum mutual information is
asymptotically approached after a certain number of generations and
the word/categories at this point typically reflect known
grammatical categories. In \citep{lin,pantel} word synonyms are
discovered in a text by taking the similarity among words based on
the mutual information between the two words and other words in the
text. Those words who have the highest similarity are often
semantically similar.

The approach in this paper most closely follows that of Lin et. al.
in finding the mutual information amongst words in the corpus and
then computing a similarity between the words based off of this. The
mutual information between two words in the text, $x$ and $y$, is
termed $I_{xy}$ and is defined as

\begin{equation}
I_{xy}=\sum_x\sum_y p_{xy}\log{\frac{p_{xy}}{p_x p_y}}
\label{mutualinfo}
\end{equation}

where $p_{xy}$ is the frequency of word pair (x, y) and $p_x$ and
$p_y$ are the frequencies of words $x$ and $y$ in the texts. Two
different arrays of mutual information were calculated for the word
distance one and two pair frequencies. Finally, a blended mutual
information is calculated using different weightings of the one and
two distance mutual information.

The blended mutual information, $I_B$, is

\begin{equation}
I_B = \sqrt{I_1^2 + (WI_2)^2}
\label{mutualinfoblend}
\end{equation}

where $I_1$ and $I_2$ are the mutual information for distance one
and two word pairs respectively and the weight, $W$, takes a value
between 0 and 1. It is difficult to find an objective value for $W$.
The method used in the paper which will be explained more in the
next section is that different values of $W$ were tested until many
known words with similar meanings had high measures of similarity.
Though this could be accused of affirming the consequent, it can be
considered a method of calibration based on our small current
knowledge.

\subsection{Step 4}

For the blended mutual information a similarity measure, $S$, was
calculated where $S$ is defined as

\begin{equation}
S_{xy} = \sum_z \frac{2I_B(x,z)I_B(y,z)}{I_B(x,z)^2 + I_B(y,z)^2}
\end{equation}
where $z$ is all words in the corpus where $z \neq x, y$.

\subsection{Step 5}

The word pairs are ranked by descending similarity and the results
analyzed. Since relatively infrequent words will likely give
spurious or insignificant results, only word pairs where both words
appeared at least three times were used in the final analysis for
comparison.

\begin{table*}[!t] \vspace{1.5ex}
{\footnotesize
\begin{tabular}{|c|c|c|c|c|c|c|c|}
\hline Rank&Word 1&Word 2&Word 1 Meaning&Word 2
Meaning&Similarity&Word 1
Count&Word 2 Count\\
\hline
 1&kdi&abr&woman&man&1&3&3\\
 \hline
2&mk&amnp&god&Amun of Napata&1&7&17\\
\hline
3&mk&kek&god&?&1&7&5\\
\hline
 4&abrsel&wwikewi&every man&?&1&3&3\\
  \hline
 5&qorte&agro&in the king?&?&0.99&3&3\\
 \hline
 6&amnp&seb&Amun of Napata&divinity??&0.98&17&15\\
 \hline
 7&qes&qor&Kush&king&0.98&12&6\\
 \hline
 8&ne&pqr&?&prince&0.98&3&3\\
\hline
9&mk&seb&god&divinity??&0.97&7&15\\
\hline
\end{tabular}}
\caption{Top word pair similarities with meanings where known}
\label{similar}
\end{table*}

In table \ref{similar}, the top word pairs by descending similarity
are shown. A similarity cutoff of 0.95 was used given the clustering
of words above 0.95 and the poor matching of known words and wider
spread of similarity scores for word pairs with a score under 0.95.
The value of $W$ used is 0.75. This value was chosen because of the
excellent and high similarity match of the first two word pairs
which consist entirely of known words with similar meanings. The
following words in the ranking also show promise. The word
\emph{kek} is still undeciphered but may likely have a religious
meaning. Perhaps 'soul' like the Egyptian \emph{ka} or a Meroitic
deity, however, this is pure speculative. The word \emph{seb} is
well-known among Meroitic scholars to have a religious meaning,
possibly the name of a deity, but the exact meaning is still
unknown. The word \emph{abrsel} means ``every man'' while though
\emph{wwikewi} isn't specifically understood, the \emph{wwi}-
morpheme indicates directional movement.

\begin{figure}
    \centering
    \includegraphics[height=7in, width=5in]{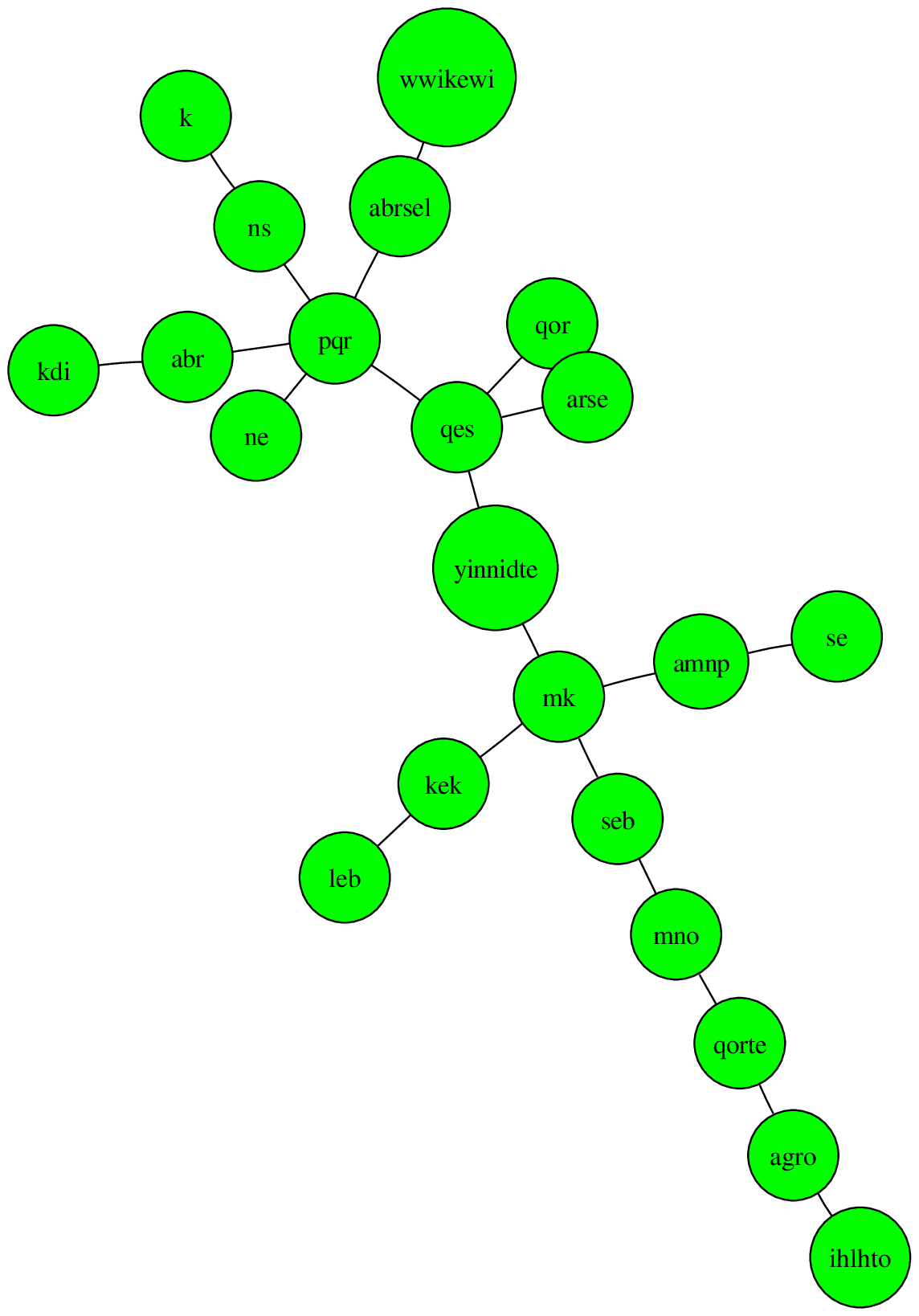}

\caption{Graphic representation of the minimum spanning tree of the
data represented in table \ref{similar}.}
\label{tree}
\end{figure}

In order to more clearly see how the words relate to each other, I
graphically visualized the similarity relationships using the
distance metric derived in \citep{gower}. This distance metric is
used to convert comparison metrics such as correlation or similarity
among variables to metric distances.

\begin{equation}
d=\sqrt{2(1-s_{ij})}
\end{equation}

where $s_{ij}$ is the similarity between words $i$ and $j$. These
distances can then be plotted onto a minimum spanning tree such as
that in figure \ref{tree}

\section{Problems \& Issues}

As stated before, I cannot claim to solve the issues related to
Meroitic solely through statistical analysis. In particular, though
the information such an analysis can provide is directional it is
sensitive to interpretation. The choice of the weight, $W$, though
not completely arbitrary uses \emph{a priori} knowledge to set its
value. While the results it returns are consistent with closely
related known words, this may introduce bias. The cutoff for the
similarity measurement, at a value of 0.95, is also arbitrary and
based on a subjective analysis of the data. Therefore, despite the
equations, much of this technique requires knowledge of the language
and subjective interpretation to extract useful knowledge. In the
end, however, I believe this technique will help shed a light on
many previously intractable problems in Meroitic and could become a
valuable tool in the eventual decipherment of the language.

\section*{Acknowledgements}

The author would like to thank Dr. Claude Rilly and Dr. Laurance
Doyle for helpful comments and critiques of the ideas in the paper.

\end{document}